\documentclass[10pt,conference]{IEEEtran}
\IEEEoverridecommandlockouts
\usepackage{cite}
\usepackage{amsmath,amssymb,amsfonts}
\usepackage{algorithmic}
\usepackage{graphicx}
\usepackage{textcomp}
\usepackage{xcolor}
\usepackage{algorithm}
\usepackage{booktabs}
\usepackage{algorithmic}
\usepackage{subcaption}
\newtheorem{theorem}{Theorem}

\newtheorem{assumption}{Assumption}
\usepackage{diagbox}
\usepackage{hyperref}
\usepackage{multirow}
\usepackage{makecell}
\def\BibTeX{{\rm B\kern-.05em{\sc i\kern-.025em b}\kern-.08em
    T\kern-.1667em\lower.7ex\hbox{E}\kern-.125emX}}
\begin{document}

\title{DFedReweighting: A Unified Framework for Objective-Oriented Reweighting in Decentralized Federated Learning
}

\author{
\IEEEauthorblockN{Kaichuang Zhang$^{1}$, Wei Yin$^{2}$, Jinghao Yang$^{3}$, Ping Xu$^{4}$\\kaichuangzhang@usf.edu, wei.yin01@utrgv.edu, jinghao.yang@utrgv.edu, pxu4@gsu.edu}
\IEEEauthorblockA{
$^{1}$Department of Electrical Engineering, University of South Florida, Tampa, Florida, USA\\
$^{2}$Department of Mathematics, The University of Texas Rio Grande Valley, Edinburg, Texas, USA\\
$^{3}$Department of Electrical and Computer Engineering, The University of Texas Rio Grande Valley, Edinburg, Texas, USA\\
$^{4}$Department of Computer Science, Georgia State University, Atlanta, Georgia, USA
}
\thanks{This work was partly supported by the National Science Foundation (Grants \#2112650, \#2434916). Corresponding author: Ping Xu (email: pxu4@gsu.edu).}
}

\maketitle
\begin{abstract}
Decentralized federated learning (DFL) has emerged as a promising paradigm that enables multiple clients to collaboratively train machine learning models through iterative rounds of local training, communication, and aggregation, without relying on a central server. Nevertheless, DFL systems continue to face a range of challenges, including fairness and Byantine robustness.
To address these challenges, we propose \textbf{DFedReweighting}, a unified aggregation framework that achieves diverse learning objectives in DFL via objective-oriented reweighting at the final step of each learning round. Specifically, for each client, the framework first evaluates a target performance metric (TPM) on a compact auxiliary dataset constructed from local data, yielding preliminary aggregation weights, which are subsequently refined by a customized reweighting strategy (CRS) to produce the final aggregation weights. Theoretically, we prove that an appropriate TPM-CRS combination guarantees linear convergence for general $L$-smoothand strongly convex functions. Empirical results consistently demonstrate that \textbf{DFedReweighting} significantly improves fairness and robustness against Byzantine attacks across diverse settings. Two multi-objective examples, spanning tasks across and within clients, further establish that a broad range of desired learning objectives can be accommodated by appropriately designing the TPM and CRS. Our code is available at \url{https://github.com/KaichuangZhang/DFedReweighting}. 
\end{abstract}


\section{Introduction}

In federated learning (FL), multiple {clients} collaboratively train a machine learning model by performing {local training} and synchronizing their {updates} (gradients or model parameters) with a {central server}. Consequently, clients maintain their raw data locally, thereby preserving data privacy~\cite{mcmahan2017communication,shokri2015privacy}.  
In contrast, decentralized federated learning (DFL), where each client independently executes both local model updates and aggregation without relying on a central server, has recently gained attention~\cite{he2021spreadgnn,lalitha2018fully}. A typical DFL system operates iteratively through a {three-stage process} comprising local training, communication, and aggregation~\cite{zhang2024Byzantine, zhang2025distributional}. This process repeats over multiple {round}s until convergence, ultimately ensuring that each client obtains an effective local model that can be deployed in real-world scenarios, such as the Internet of Things~\cite{chellapandi2023federated}, healthcare~\cite{baghersalimi2023decentralized}, etc.
\label{sec:intro}
\begin{figure}[t]
\begin{center}
\centering
\includegraphics[width=0.45\textwidth]{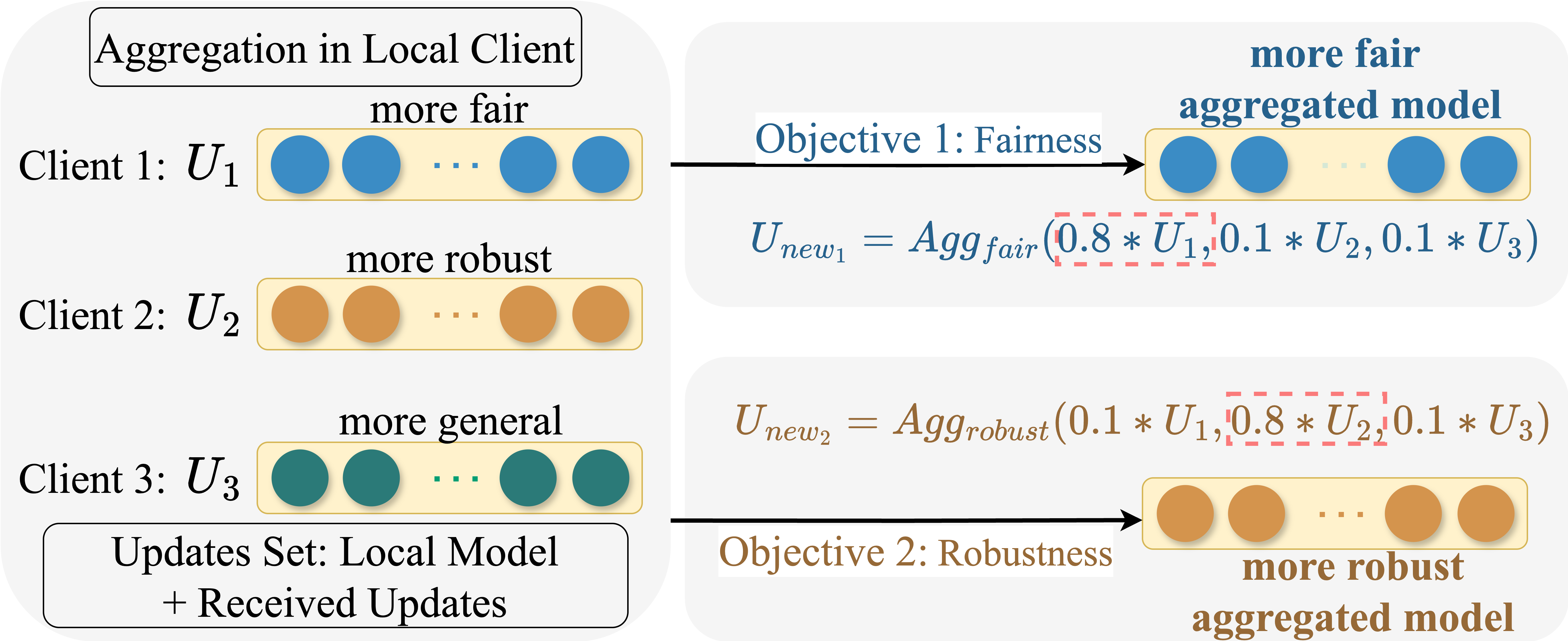}
\end{center}
\caption{\textit{Objective-oriented reweighting aggregation.} Each client assigns aggregation weights according to its local objective. For instance, a client seeking to improve fairness may assign a higher weight to update $U_1$ if it yields superior fairness performance. More generally, each client allocates weights to received updates ($U_1, U_2, U_3$) through the reweighting aggregation function $Agg ( \,)$, enabling diverse objectives to be pursued simultaneously across clients.
}
\label{fig:motivation}
\end{figure}

Although DFL mitigates issues inherent to centralized FL, such as single point of failure~\cite{chen2021robust} and communication bottlenecks~\cite{le2024exploring}, other critical challenges, e.g., robustness, fairness, communication efficiency, etc., persist in DFL~\cite{yuan2024decentralized}.
Addressing these challenges typically involves developing dedicated algorithms at any stage of the three-stage (i.e., local computation, communication, local aggregation) process in DFL. However, we argue that the final aggregation stage, characterized by {target performance metrics} (TPMs) and {customized reweighting strategies} (CRSs), is sufficient to resolve these issues. In essence, numerous DFL methods ultimately take effect at the final aggregation stage regardless of the specific problems they aim to address.
For example, \texttt{q-FFL}~\cite{li2019fair} promotes fairness by modifying the global objective function, a change that directly affects local updates while implicitly inducing reweighted aggregation. Similarly, \texttt{FedFV}~\cite{wang2021federated} alleviates gradient conflicts through iterative adjustments to both the direction and magnitude of gradients, which constitutes another indirect form of reweighted aggregation.

In contrast to these inherently indirect methods, directly applying objective-oriented reweighting on updates (as shown in Fig. \ref{fig:motivation}) at the final aggregation stage provides a more generalizable and transparent approach to steer the learning process. Rather than relying on implicit or heuristic modifications, this approach explicitly aligns aggregation with the underlying optimization objective — intuitively, when heterogeneous updates are available, effective learning demands prioritizing those most relevant to the target objective. The objective-oriented reweighting thus provides a principled aggregation strategy for DFL systems to distinguish useful received updates from noisy or irrelevant ones, since these updates may differ substantially due to data heterogeneity, varying objectives, or unreliable participants.  Specifically, objective-oriented reweighting enables each client to treat received updates as heterogeneous sources, selectively emphasizing those that best support its local objective before integration. The resulting aggregation scheme balances adaptability and robustness while preserving the decentralized nature of the system.

The contributions of this work are summarized as follows:
\begin{itemize}
   \item We propose \textbf{DFedReweighting}, a unified aggregation framework that addresses a broad range of challenges in DFL via objective-oriented reweighting at the final step of each learning round.
   \item We provide a theoretical convergence analysis establishing that any appropriate TPM-CRS combination guarantees linear convergence for general $L$-smooth and strongly convex objective functions.
  \item We conduct extensive experiments demonstrating that \textbf{DFedReweighting} consistently achieves superior performance on client fairness and Byzantine robustness across diverse settings. Additionally, two multi-objective examples illustrate the flexibility of the TPM-CRS design paradigm in accommodating a wider range of learning objectives.

\end{itemize}
\section{Related Work}
This section reviews representative works addressing key challenges in DFL systems across different stages of the learning process. Given the structural similarities between FL and DFL, client-side methods originally developed for FL are also discussed where relevant.


\textbf{Local Training Stage}: 
Some studies pursue specific objectives, notably fairness, by modifying the local objective function or altering the training workflow. AFL~\cite{mohri2019afl} promotes fairness by minimizing the worst-case loss across all clients. FedMGDA+~\cite{hu2020fedmgda+} enhances both fairness and robustness through multi-objective optimization over clients' local losses. GIFAIR-FL~\cite{yue2023gifair} introduces a regularization term that penalizes dispersion in per-group losses, steering optimization toward fairer solutions. FairFed~\cite{Ezzeldin2021FairFed} enforces per-group loss upper bounds via server-side gradient reweighting, improving global fairness without sacrificing accuracy. AgnosticFair~\cite{du2021fairness} assigns sample-level reweighting values through kernel functions applied to both the loss and the fairness constraint.

\textbf{Communication Stage}:
A separate line of work targets communication efficiency. \cite{konevcny2016federated} propose structured and sketched updates to reduce uplink communication costs by up to two orders of magnitude. \cite{caldas2018expanding} leverage lossy compression and dropout to further reduce communication overhead. FedKD~\cite{wu2022communication} achieves communication efficiency through adaptive mutual knowledge distillation combined with dynamic gradient compression. \cite{ribero2024reducing} improve efficiency by selectively soliciting updates only from clients with informative gradients while estimating the remainder. \cite{shah2021model} investigate structured sparsification techniques to construct compact models with substantially reduced storage and bandwidth requirements.
 
\textbf{Aggregation Stage}:
A third category of methods pursues their objectives at the aggregation stage. For fairness, pFedFair~\cite{lei2025pfedfair} identifies under performing clients, assigns them higher fairness coefficients during aggregation, and subsequently personalizes local fine-tuning, achieving a Pareto-optimal fairness-accuracy trade-off. For robustness, \cite{li2020learning} propose a spectral anomaly detection model based on a variational autoencoder with dynamic thresholds, deployed at the central server to identify and suppress malicious updates. FLTrust~\cite{cao2020fltrust} computes per-worker trust scores using a curated root dataset at the server to filter unreliable updates before aggregation. Siren~\cite{guo2021siren} combines a client-side alarming mechanism with a server-side root-dataset-based defense to identify malicious participants. ByGARS~\cite{regatti2020bygars} derives reputation scores for each worker using a server-held auxiliary dataset and incorporates them into adaptive gradient aggregation.
\section{Problem Statement}

\label{sec:problem_statement}

A DFL system can be modeled as a decentralized network of $N$ clients interconnected over a fixed undirected graph $\mathcal{G} = (\mathcal{N}, \mathcal{E})$, where $\mathcal{N}$ denotes the client set and $\mathcal{E}$ denotes the edge set~\cite{zhang2024Byzantine}. Any two clients $n$ and $m$ are said to be one-hop neighbors if $(n, m) \in \mathcal{E}$, enabling them to exchange information during communication steps. For client $k$, its one-hop neighbors are in the set $\mathcal{N}_k = \{m|(m, k) \in \mathcal{E}\}$. Under the assumption that all clients work properly, the learning task of a DFL system is to minimize the weighted average of local clients' losses, given by

\begin{equation}
\label{eq:dfl_problem}
    \min_{\mathbf{W} } F(\mathbf{w}) = \sum_{k=1}^{N} p_k f_k(\mathbf{w}_k, \mathcal{D}_k),
\end{equation}
where $\mathbf{W} := \{\mathbf{w}_1, \mathbf{w}_2, \dots, \mathbf{w}_{N}\}$ denotes the set of all local models, for which a consensus constraint may or may not be imposed depending on the task. $\mathcal{D}_k$ represents client $k$'s local data, and $\mathcal{D} = \bigcup_{k \in \mathcal{N}} \mathcal{D}_k$. $p_k$ is the aggregation weight assigned to client $k$, which is set to be $p_k = \frac{|\mathcal{D}_k|}{|\mathcal{D}|}$ in DFedAvg~\cite{sun2022decentralized}, with $|\mathcal{D}_k|$ representing the number of local data samples owned by client $k$.


Various popular decentralized algorithms, such as decentralized stochastic gradient descent and the decentralized alternating direction method of multipliers, can be employed to solve~\eqref{eq:dfl_problem}. Taking decentralized SGD as an example, at each communication round $t$, each client $k$ $(k \in \mathcal{N})$ executes the following three-stage learning process multiple times until convergence~\cite{fang2024byzantine, wu2023byzantine}.

\textbf{Stage 1 (local training):} client $k$ takes a local SGD step to obtain an intermediate update $\mathbf{w}_k^{t+\frac{1}{2}}$:
\begin{equation}
\label{eq:local_update_SGD}
\mathbf{w}_k^{t+\frac{1}{2}} = \mathbf{w}_k^{t} - \eta^t \nabla f_k(\mathbf{w}_k^{t}; \xi^{t}_k),
\end{equation}
where $\mathbf{w}_k^{t+\frac{1}{2}}$ is the locally trained model at iteration $t$, $\eta^t$ is the learning rate, and $\nabla f_k(\mathbf{w}_k^{t}; \xi^{t}_k)$ represents the stochastic gradient of the local objective function $f_k$ for client $k$, evaluated using the model parameter $\mathbf{w}_k^t$ on a mini-batch sample $\xi_k^t$ drawn from the client’s local dataset $\mathcal{D}_k$. 

\textbf{Stage 2 (communication):} client $k$ sends $\mathbf{w}_k^{t+\frac{1}{2}}$ to its neighbors $m\;(m\in \mathcal{N}_k)$ and receives $\mathbf{w}_m^{t+\frac{1}{2}}$ from its neighbors, resulting in a local update set $\mathbf{W}_k^t$ that includes client $k$'s local model and updates from its neighbors:
\begin{equation}
\label{eq:communication}
\mathbf{W}_k^t := \{ \mathbf{w}_k^{t+\frac{1}{2}} \} \cup \{ \mathbf{w}_m^{t+\frac{1}{2}} | m \in \mathcal{N}_k\}.
\end{equation}

\textbf{Stage 3 (aggregation):} client $k$ follows~\eqref{equ:local_aggregation} to get a new local model \cite{sun2022decentralized}:
\begin{equation}
\label{equ:local_aggregation}
\mathbf{w}_k^{t+1} = \sum_{\mathbf{w}_i^{t+\frac{1}{2}} \in \mathbf{W}_k^t} p_i \mathbf{w}_i^{t+\frac{1}{2}},
\end{equation}
where $\mathbf{w}_k^{t+1}$ is the new local model for client $k$ at round $t+1$.

The DFL system repeats this 3-stage learning process until all local models converge, at which point they can be deployed for practical use.
\section{DFedReweighting}


\begin{figure*}[t]
\begin{center}
\begin{minipage}{1\linewidth}
\centering
\includegraphics[width=.80\textwidth]{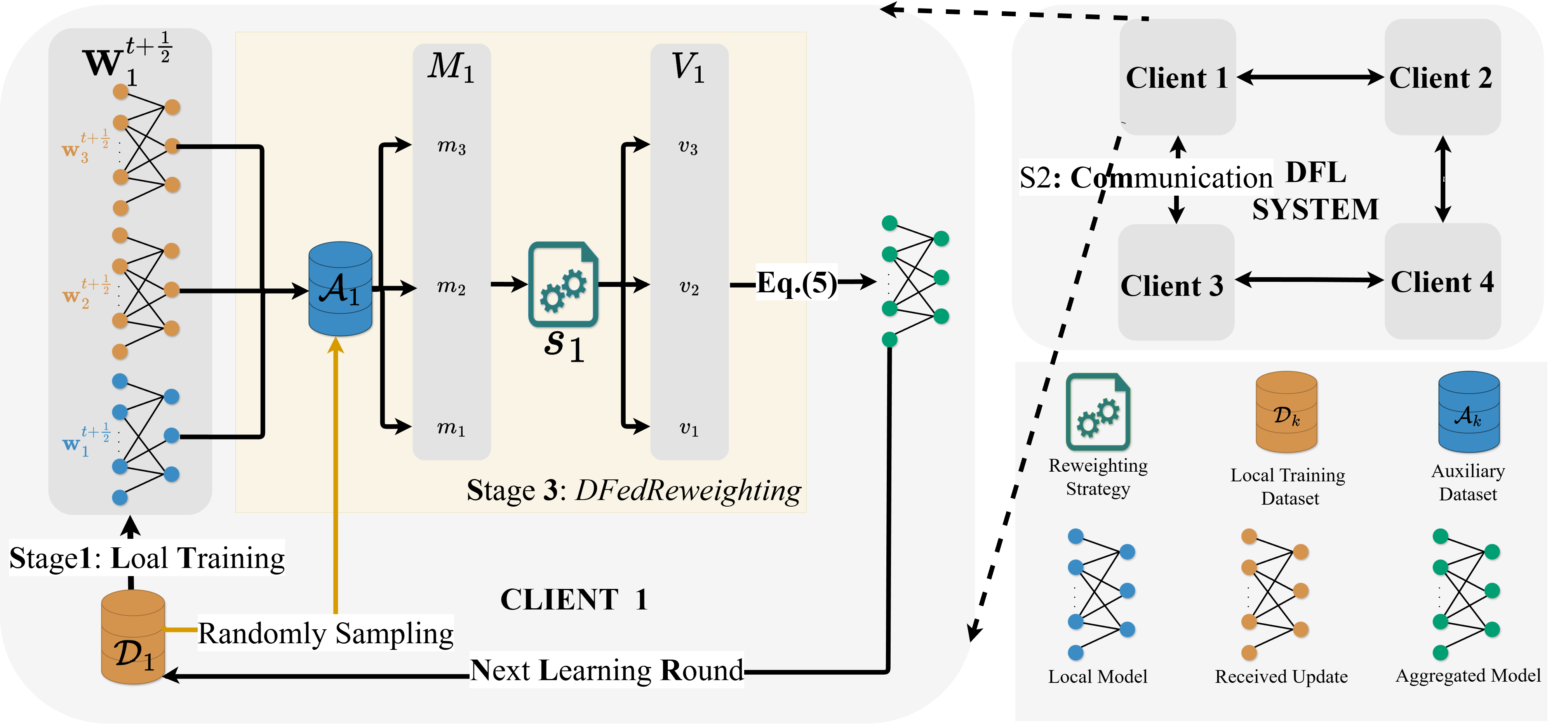}
\end{minipage}
\end{center}
\caption{\textit{Overview of \textbf{DFedReweighting} in a four-client DFL system (each client has two neighbors).} Each client proceeds in four steps: (i) constructs an auxiliary dataset $\mathcal{A}_k$ from local data; (ii) evaluate the TPM on $\mathcal{A}_k$ to obtain a preliminary performance score set $M_k$; (iii) apply the CRS to $M_k$ to derive the final aggregation weight set $V_k$; and (iv) aggregates models using $V_k$. 
}
\label{fig:overview}
\end{figure*}

We propose \textbf{DFedReweighting}, a unified aggregation framework (Algorithm~\ref{framework}) to address key challenges in DFL systems through objective-oriented reweighting at the final stage of each learning round. The core insight is that each client can steer its own learning behavior by assigning aggregation weights that reflect its local objective, rather than relying on default, objective-agnostic schemes. Specifically, for each client kk
k, the framework first evaluates a \textit{target performance metric} (TPM) on a compact auxiliary dataset constructed from local data, yielding preliminary aggregation weights. These weights are subsequently refined by a \textit{customized reweighting strategy} (CRS) to produce the final aggregation weights. The framework proceeds in four steps:

\begin{algorithm}[t]
\caption{DFedReweighting runs at client $k$.}
\textbf{Input:} $\mathbf{W}_k^t = \{\mathbf{w}_k^{t+\frac{1}{2}}\} \cup \{\mathbf{w}_m^{t+\frac{1}{2}} \mid m \in \mathcal{N}_k\}$, local training dataset $\mathcal{D}_k$, sampling ratio $\alpha$, reweighting strategy $s$. \\
\textbf{Output:} $\mathbf{w}_k^{t+1}$

\begin{algorithmic}[1]
\STATE Subsample a fraction $\alpha$ of $\mathcal{D}_k$ to form auxiliary dataset $\mathcal{A}_k$.
\FOR{each $\mathbf{w}_i^{t+\frac{1}{2}} \in \mathbf{W}_k^t$}
    \STATE Calculate the performance score $m_i$ for update $\mathbf{w}_k^{t+\frac{1}{2}}$ using $\mathcal{A}_k$ to form $M_k$.
\ENDFOR
\FOR{each $m_i\in M_k$}
    \STATE Compute the corresponding final aggregation weight $v_i^t = s(m_i, M_k)$ based on performance score $m_o$ and reweighting strategy $s$ to form $V_k$.
\ENDFOR
\STATE Update local model: $\mathbf{w}_k^{t+1} = \sum_{\mathbf{w}_i^{t+\frac{1}{2}} \in \mathbf{W}_k^t} v_i^t  \mathbf{w}_i^{t+\frac{1}{2}}$.
\STATE \textbf{Return} $\mathbf{w}_k^{t+1}$.
\end{algorithmic}
\label{framework}
\end{algorithm}

\textbf{Step 1 (Construct Auxiliary Dataset $\mathcal{A}_k$)}  
%
Unlike conventional FL, where a central server aggregates client updates without access to any local data~\cite{abyane2022mda}, DFL enables each client to validate received updates against its own local training data, a capability that is fundamental to the objective-oriented reweighting framework proposed in this work. Nevertheless, utilizing the entire local dataset for this validation process introduces unnecessary computational overhead. To address this, we construct a compact auxiliary dataset $\mathcal{A}_k$ by \textit{subsampling} a representative subset of the local training data, which is sufficient to serve as the basis for update evaluation while remaining computationally efficient.

\textbf{Step 2 (Obtain Preliminary Weights $M_k$)}. 
A key advantage of DFL is that each client retains full autonomy over its local aggregation process, enabling the incorporation of objective-aware criteria directly into the update selection mechanism. Leveraging this flexibility, the proposed framework computes preliminary aggregation weights based on a TPM evaluated on the auxiliary dataset $\mathcal{A}_k$. Specifically, upon constructing $\mathcal{A}_k$in Step 1, client $k$ evaluates the TPM against all received model updates $\mathbf{W}_k^t$ as well as its own locally trained model, producing a performance score set $M_k$. The TPM can be instantiated as accuracy, loss, or any other objective-specific measure, providing the framework with the generality to accommodate diverse learning goals.

\textbf{Step 3 (Calculate Final Aggregation Weights $V_k$).} 
Reweighting-based aggregation offers a principled solution, enabling each client in a DFL system to selectively prioritize updates from neighboring peers according to its own objective, rather than relying on default weights derived from dataset sizes. Specifically, after obtaining $M_k$ the framework applies a CRS $s(m_i, M_k)$ to produce a new weight set $V_k$ tailored to the client's goal — for instance, upweighting the local model to promote personalization, or assigning near-zero weights to suspected Byzantine participants to enhance robustness. The flexibility of this design lies in the choice of $s$ by appropriately pairing the TPM with a suitable CRS, the framework can accommodate a broad range of learning objectives, as demonstrated through concrete TPM-CRS design examples in the Experiments section.

\textbf{Step 4 (Aggregation.)}  After obtaining final aggregation weights $V_k$ from Step 3, each client performs aggregation using
\begin{equation}
\label{equ:agg}
\mathbf{w}_k^{t+1} = \sum_{\mathbf{w}_i^{t+\frac{1}{2}} \in \mathbf{W}_k^t} v_i^t  \mathbf{w}_i^{t+\frac{1}{2}}.
\end{equation}

Fig.~\ref{fig:overview} shows an example of DFedReweighting in a ring-structured DFL system with four clients. All clients follow a three-stage learning process. 
\section{Convergence Analysis}
Building upon the definitions and notations introduced in Section~\ref{sec:problem_statement}, we analyze the convergence properties of \textbf{DFedReweighting}. Since each client independently maintains its own local model in DFL, our analysis focuses on the convergence behavior of each client's local model $\mathbf{w}_k$. The analysis adopts the following standard assumptions.

\begin{assumption}{$L$-smoothness~\cite{ma2022shieldfl,wu2023byzantine,ye2022decentralized}.}
\label{assumption:L_smooth}
Each local objective function $f_k$ has $L$-Lipschitz continuous gradients, i.e., for all $\mathbf{w}, \mathbf{w}' \in \mathbb{R}^d$,
$ \|\nabla f_k(\mathbf{w}) - \nabla f_k(\mathbf{w}')\| \leq L \|\mathbf{w} - \mathbf{w}'\|$. 
\end{assumption}

\begin{assumption}{Strong convexity~\cite{jiang2025performanceanalysisdecentralizedfederated,li2023convergence}.} All local objective functions $\{f_k, k \in \mathcal{N}\}$ are strongly convex.
\end{assumption}

Additional standard assumptions, including bounded second moment \cite{ma2022shieldfl,wu2023byzantine,ye2022decentralized}, connected graph $\mathcal{G}_\mathcal{B}$~\cite{fang2024byzantine,yang2024byzantine}, and unbiased stochastic gradients with bounded variance~\cite{ye2022decentralized}, are adopted without restatement.

In \textbf{DFedReweighting}, the aggregation weights satisfy $v_i^t \ge 0$ for any TPM-CRS combination, as defined in~\eqref{equ:agg}.  
For any pair $(k, i) \notin \mathcal{E}$, i.e., when no direct connection exists between clients $k$ and $i$, we set $v_i^t=0$.

Our analysis begins by applying the descent lemma~\cite{jiang2025performanceanalysisdecentralizedfederated,li2023convergence} for $L$-smooth, strongly convex functions: 
\begin{equation}
    \label{eq:lemma}
    \mathbb{E}\left[F(\mathbf{w}_k^{t+1})\right] - F(\mathbf{w}^*_k) \leq \frac{L}{2} \cdot \mathbb{E}\left[\|\mathbf{w}_k^{t+1} - \mathbf{w}^*_k\|^2\right],
\end{equation}
where $\mathbf w_k^*$ is a deterministic optimal point for client $k$.

Substituting the aggregation rule \eqref{equ:agg} and the local update rule \eqref{eq:local_update_SGD} into \eqref{eq:lemma}, the squared error decomposes into three terms:
\begin{equation}
\label{equ:3terms_inmain}
\begin{aligned}
\bigl\|\mathbf w_k^{t+1}-\mathbf w_k^{*}\bigr\|^{2}
&=\Bigl\|\sum_{i=1}^{N}v_i^{t}(\mathbf w_i^{t}-\mathbf w_k^{*})
           -\eta^{t}\sum_{i=1}^{N}v_i^{t}\nabla f_i(\mathbf w_i^{t};\xi_i^{t})\Bigr\|^{2}\\
&=\underbrace{\Bigl\|\sum_{i=1}^{N}v_i^{t}(\mathbf w_i^{t}-\mathbf w_k^{*})\Bigr\|^{2}}_{\text{Term 1}} \\
&\quad+\;
\underbrace{(\eta^{t})^{2}\Bigl\|\sum_{i=1}^{N}v_i^{t}\nabla f_i(\mathbf w_i^{t};\xi_i^{t})\Bigr\|^{2}}_{\text{Term 2}}\\
&\quad-\;
\underbrace{2\eta^{t}\bigl\langle\sum_{i=1}^{N}v_i^{t}(\mathbf w_i^{t}-\mathbf w_k^{*}),
                                  \sum_{i=1}^{N}v_i^{t}\nabla f_i(\mathbf w_i^{t};\xi_i^{t})\bigr\rangle}_{\text{Term 3}}\!.
\end{aligned}
\end{equation}

Each term is bounded as follows. For Term 1, by Jensen's inequality:
\begin{equation}
\label{equ:term1_inmain}
\begin{aligned}
\left\| \sum_{i=1}^N v_i^t (\textbf{w}_i^t - \textbf{w}_k^*) \right\|^2 
\le \sum_i v_i^{t}\bigl\lVert \textbf{w}_i^{t}-\textbf{w}_k^\star\bigr\rVert^{2}.
\end{aligned}
\end{equation}

For Term 2, using the bounded gradient assumption:
\begin{equation}
\label{equ:term2_inmain}
    (\eta^t)^2 \left\| \sum_{i=1}^N v_i^t \nabla f_i(\textbf{w}_i^t; \xi_i^{t}) \right\|^2 
\leq (\eta^t)^2 G^2 \Rightarrow (\eta^t \cdot G)^2.
\end{equation}
For Term 3: 
\begin{equation}
\label{equ:term3_inmain}
\begin{aligned}
    \left\langle \sum_{i=1}^{N} v_i^t (\textbf{w}_i^t - \textbf{w}_k^*), \sum_{i=1}^{N} v_i^t \nabla f_i(\textbf{w}_i^t; \xi_i^{t}) \right\rangle &\quad \\
    \leq \frac{3L}{2}\sum_{i=1}^{N} v_i^t \|\textbf{w}_i^t - \textbf{w}_k^*\|^2.
\end{aligned}
\end{equation}

Combining these bounds establishes the following convergence theorem.
\begin{theorem}
For any $L$-smooth, strongly convex local objective function at client $k$, \textbf{DFedReweighting} with a {dynamic} learning rate $\eta^t < 1$  satisfies the following recursive bound:
\begin{equation}
    \label{eq:recursive_bound_inmain}
    \begin{aligned}
    \lVert\mathbf w_k^{t+1}-\mathbf w_k^{*}\rVert^{2}
    &\le
    \Bigl(\prod_{h=0}^{t}\bigl(1-3\eta^{h}L\bigr)\Bigr)
    \Bigl(\sum_{i=1}^{N}v_i^{0}\,\lVert\mathbf w_i^{0}-\mathbf w_k^{*}\rVert^{2}\Bigr)\\
    &\quad+\;
    \sum_{j=0}^{t}
    \Bigl(\prod_{h=j+1}^{t}\bigl(1-3\eta^{h}L\bigr)\Bigr)
    \bigl(\eta^{j}G\bigr)^{2}.
    \end{aligned}
    \end{equation}
\end{theorem}
In particular, setting $\eta^t < 1/(3L)$ at all rounds ensures that the first term contracts geometrically, and the expected squared error converges linearly to a neighborhood of order   $\eta G^2 / L$.

\section{Experiments}
\begin{table*}[th]
    \centering
    \textbf{(a) MNIST.} \\
    \resizebox{1.0\textwidth}{!}{
    \begin{tabular}{c|c|c|c|c|c|c|c}
        \toprule[1pt]
        \diagbox{\textbf{DH}}{\textbf{Methods}} 
        &\textbf{DFedAvg}&\textbf{\makecell[c]{$q$-FDFL\\($q=0.1$)}}&\textbf{\makecell[c]{$q$-FDFL\\($q=0.01$)}}&\textbf{\makecell[c]{$q$-FDFL\\($q=0.5$)}}&\textbf{\makecell[c]{DFedReweighting\\($\mathcal{T}=0.01$)}}&\textbf{\makecell[c]{DFedReweighting\\($\mathcal{T}=0.1$)}}&\textbf{\makecell[c]{DFedReweighting\\($\mathcal{T}=0.5$)}}\\ \toprule[1pt]
    IID       &0.807{\footnotesize(91.867)}&0.964{\footnotesize(91.760)}&0.787{\footnotesize(91.873)}&\textbf{0.608}{\footnotesize(91.254)}&0.896{\footnotesize(91.846)}&0.850{\footnotesize(92.120)}&0.838{\footnotesize(91.996)}\\ \toprule[1pt]
Diri(0.1)   &380.611{\footnotesize(79.454)}&95.408{\footnotesize(84.953)}&110.003{\footnotesize(84.573)}&29.335{\footnotesize(88.512)}&9.191{\footnotesize(95.994)}&\textbf{8.731}{\footnotesize(95.900)}&56.058{\footnotesize(92.576)}\\ \toprule[1pt]
Diri(1.0)   &9.712{\footnotesize(91.339)}&8.030{\footnotesize(91.628)}&8.023{\footnotesize(91.656)}&8.539{\footnotesize(91.457)}&9.383{\footnotesize(90.941)}&\textbf{5.870}{\footnotesize(92.512)}&7.099{\footnotesize(92.006)}\\ \toprule[1pt]
LabelSkew(4)&107.049{\footnotesize(85.449)}&220.539{\footnotesize(80.629)}&188.987{\footnotesize(79.877)}&13.646{\footnotesize(91.036)}&5.725{\footnotesize(95.598)}&\textbf{4.125}{\footnotesize(95.414)}&6.468{\footnotesize(95.328)}\\ \toprule[1pt]
    \end{tabular}
    }

    \textbf{(b) Fashion MNIST.} \\
    \resizebox{1.0\textwidth}{!}{
    \begin{tabular}{c|c|c|c|c|c|c|c}
        \toprule[1pt]
        \diagbox{\textbf{DH}}{\textbf{Methods}} 
        &\textbf{DFedAvg}&\textbf{\makecell[c]{$q$-FDFL\\($q=0.1$)}}&\textbf{\makecell[c]{$q$-FDFL\\($q=0.01$)}}&\textbf{\makecell[c]{$q$-FDFL\\($q=0.5$)}}&\textbf{\makecell[c]{DFedReweighting\\($\mathcal{T}=0.01$)}}&\textbf{\makecell[c]{DFedReweighting\\($\mathcal{T}=0.1$)}}&\textbf{\makecell[c]{DFedReweighting\\($\mathcal{T}=0.5$)}}\\ \toprule[1pt]
IID       &1.455{\footnotesize(81.597)}&1.237{\footnotesize(83.040)}&1.151{\footnotesize(81.757)}&1.153{\footnotesize(83.947)}&\textbf{1.005}{\footnotesize(82.460)}&1.243{\footnotesize(82.147)}&1.196{\footnotesize(81.653)}\\ \toprule[1pt]
Diri(0.1)  &360.772{\footnotesize(76.972)}&335.066{\footnotesize(76.112)}&429.660{\footnotesize(76.493)}&414.974{\footnotesize(74.970)}&16.657{\footnotesize(95.881)}&\textbf{13.422}{\footnotesize(95.390)}&161.933{\footnotesize(86.862)}\\ \toprule[1pt]
Diri(1.0)   &22.610{\footnotesize(83.172)}&18.020{\footnotesize(83.204)}&18.012{\footnotesize(82.923)}&20.125{\footnotesize(83.686)}&14.253{\footnotesize(84.003)}&\textbf{10.452}{\footnotesize(85.549)}&11.976{\footnotesize(83.995)}\\ \toprule[1pt]
LabelSkew(4)&184.043{\footnotesize(75.315)}&199.828{\footnotesize(73.005)}&147.575{\footnotesize(72.308)}&215.891{\footnotesize(74.968)}&17.589{\footnotesize(91.773)}&\textbf{15.876}{\footnotesize(92.182)}&35.302{\footnotesize(89.697)}\\ \toprule[1pt]
    \end{tabular}
    }
\caption{\textit{Client fairness comparison (Var(Acc)) on MNIST and Fashion-MNIST.} For \textbf{DFedReweighting}, results are obtained using the TPM-CRS combination defined in \eqref{equ:combination-fairness-acc}. \textbf{Boldface} denotes the lowest variance within each scenario.}
    \label{tab:client_fairness-comparison}
\end{table*}
To validate the effectiveness and broad applicability of \textbf{DFedReweighting}, we conduct simulation experiments targeting two representative challenges in DFL: client fairness and Byzantine robustness. Beyond these primary evaluations, we further demonstrate the framework's versatility through multi-objective tasks, both across and within clients, illustrating how the TPM-CRS design can be readily adapted to address a wider range of learning objectives in DFL.

\subsection{Experimental Settings}
We begin by summarizing the key experimental settings adopted throughout this work.

\textbf{\textit{Network Topologies.}} We model the communication network as a random Erdős–Rényi graph with $N = |\mathcal{N}|$ nodes and a connection probability $\rho$.  
For Byzantine attack scenarios, following \cite{zhang2024Byzantine,ye2023tradeoff}, data samples are allocated exclusively to benign nodes while malicious nodes hold no data, with $|\mathcal{N}| = |\mathcal{B}| + |\mathcal{M}|$ denoting the total, benign, and malicious node counts, respectively. Unless otherwise specified, we set $|\mathcal{B}| = 10$, $|\mathcal{M}| = 2$, and $\rho = 0.7$.

\textbf{\textit{Data Heterogeneity (DH).}}  
Data heterogeneity is a critical factor in DFL, spanning a spectrum from IID to non-IID distributions. In the IID setting, each client's local dataset contains all classes with equal sample counts. For non-IID settings, we adopt two strategies: LabelSkew ($h$)~\cite{fang2020local}, which assigns $h$ classes to each node with equal samples per class, and Diri ($\alpha$)\cite{guo2023fedgr}, which allocates samples across clients according to a Dirichlet distribution parameterized by $\alpha$.

\textbf{\textit{Evaluation Metric.}} For classification tasks, we report the average test accuracy across all clients, $ \mathrm{Acc} = \frac{1}{N}\sum_{k=1}^{N} \mathrm{Acc}_{k} $,
where $\mathrm{Acc}_k$ is the test accuracy of client $k$ on its local or global test dataset, depending on the task. All results are averaged over four runs with different random seeds ($seed = 43, 44, 45, 46$).
\begin{table*}[th]
    \centering
    \textbf{(a) MNIST.} \\
    \resizebox{0.90\textwidth}{!}{
    \begin{tabular}{c|c|c|c|c|c|c|c|c}
        \toprule[1pt]
        \textbf{Attack} & \textbf{DH} & \textbf{No-AT} & \textbf{Median} &\textbf{mKrum} & \textbf{TM} & \textbf{Flame} & \textbf{BALANCE}  &\textbf{Ours} \\
    \toprule[1pt]
    \multirow{4}{*}{GA} 
&IID       &92.30{\scriptsize±0.06 }&91.79{\scriptsize±0.20 }&89.43{\scriptsize±1.97 }&\textbf{92.39}{\scriptsize±0.07 }&\textbf{\textcolor{black!40}{91.98}}{\scriptsize±0.66 }&91.19{\scriptsize±0.87 }&91.44{\scriptsize±0.92 }\\ \cline{2-9}
&Diri(0.1)   &79.67{\scriptsize±4.26 }&68.18{\scriptsize±5.82 }&62.57{\scriptsize±5.35 }&\textbf{\textcolor{black!40}{77.85}}{\scriptsize±3.32 }&68.26{\scriptsize±17.99 }&\textbf{79.50}{\scriptsize±1.63 }&76.40{\scriptsize±0.87 }\\ \cline{2-9}
&Diri(1.0)   &91.04{\scriptsize±0.31 }&90.50{\scriptsize±0.54 }&86.69{\scriptsize±1.44 }&\textbf{\textcolor{black!40}{90.53}}{\scriptsize±0.53 }&90.47{\scriptsize±0.40 }&\textbf{90.99}{\scriptsize±0.49 }&89.98{\scriptsize±0.16 }\\ \cline{2-9}
&LabelSkew(4)&81.32{\scriptsize±0.30 }&71.55{\scriptsize±0.43 }&61.42{\scriptsize±0.00 }&70.42{\scriptsize±3.29 }&68.49{\scriptsize±1.41 }&\textbf{\textcolor{black!40}{79.95}}{\scriptsize±0.77 }&\textbf{80.55}{\scriptsize±2.18 }\\ \cline{2-9}
 \toprule[1pt]
 \multirow{4}{*}{S-F} 
&IID       &92.30{\scriptsize±0.06 }&90.56{\scriptsize±0.56 }&91.40{\scriptsize±0.00 }&91.29{\scriptsize±0.00 }&\textbf{\textcolor{black!40}{92.35}}{\scriptsize±0.10 }&91.19{\scriptsize±0.87 }&\textbf{92.38}{\scriptsize±0.14 }\\ \cline{2-9}
&Diri(0.1)   &79.67{\scriptsize±4.26 }&51.27{\scriptsize±13.16 }&78.79{\scriptsize±0.00 }&72.69{\scriptsize±9.27 }&\textbf{\textcolor{black!40}{87.35}}{\scriptsize±2.23 }&79.50{\scriptsize±1.63 }&\textbf{88.37}{\scriptsize±0.14 }\\ \cline{2-9}
&Diri(1.0)   &91.04{\scriptsize±0.31 }&90.00{\scriptsize±0.60 }&88.99{\scriptsize±0.00 }&89.65{\scriptsize±0.68 }&\textbf{91.26}{\scriptsize±0.14 }&\textbf{\textcolor{black!40}{90.99}}{\scriptsize±0.49 }&90.45{\scriptsize±1.13 }\\ \cline{2-9}
&LabelSkew(4)&81.32{\scriptsize±0.30 }&65.49{\scriptsize±11.64 }&\textbf{89.55}{\scriptsize±0.10 }&84.06{\scriptsize±0.00 }&84.96{\scriptsize±0.72 }&79.95{\scriptsize±0.77 }&\textbf{\textcolor{black!40}{89.21}}{\scriptsize±0.40 }\\ \cline{2-9}
 \toprule[1pt]
 \multirow{4}{*}{ALIE} 
&IID       &92.30{\scriptsize±0.06 }&91.29{\scriptsize±0.58 }&90.42{\scriptsize±1.56 }&\textbf{92.37}{\scriptsize±0.08 }&\textbf{\textcolor{black!40}{92.33}}{\scriptsize±0.11 }&91.19{\scriptsize±0.87 }&92.18{\scriptsize±0.01 }\\ \cline{2-9}
&Diri(0.1)   &79.67{\scriptsize±4.26 }&83.47{\scriptsize±1.88 }&69.29{\scriptsize±7.10 }&\textbf{\textcolor{black!40}{83.55}}{\scriptsize±2.96 }&\textbf{87.84}{\scriptsize±1.84 }&82.41{\scriptsize±0.54 }&73.21{\scriptsize±2.06 }\\ \cline{2-9}
&Diri(1.0)   &91.04{\scriptsize±0.31 }&90.45{\scriptsize±0.23 }&87.57{\scriptsize±1.01 }&90.46{\scriptsize±0.58 }&\textbf{91.26}{\scriptsize±0.16 }&\textbf{\textcolor{black!40}{90.99}}{\scriptsize±0.49 }&90.66{\scriptsize±0.82 }\\ \cline{2-9}
&LabelSkew(4)&81.32{\scriptsize±0.30 }&81.65{\scriptsize±1.58 }&\textbf{89.55}{\scriptsize±0.10 }&87.36{\scriptsize±1.52 }&86.30{\scriptsize±0.64 }&79.95{\scriptsize±0.77 }&\textbf{\textcolor{black!40}{88.90}}{\scriptsize±0.28 }\\ \cline{2-9}
 \toprule[1pt]

    \end{tabular}
    }
    
    \textbf{(b) Fashion MNIST.} \\
    \resizebox{0.90\textwidth}{!}{
    \begin{tabular}{c|c|c|c|c|c|c|c|c}
        \toprule[1pt]
        \textbf{Attack} & \textbf{DH} & \textbf{No-AT} & \textbf{Median} &\textbf{mKrum} & \textbf{TM}& \textbf{Flame}& \textbf{BALANCE}  & \textbf{Ours} \\
         \toprule[1pt] \multirow{4}{*}{GA} 
&IID       &81.60{\scriptsize±0.68 }&81.52{\scriptsize±0.33 }&81.28{\scriptsize±0.86 }&81.21{\scriptsize±0.80 }&\textbf{81.87}{\scriptsize±0.52 }&\textbf{\textcolor{black!40}{81.70}}{\scriptsize±0.47 }&81.32{\scriptsize±0.52 }\\ \cline{2-9}
&Diri(0.1)   &71.82{\scriptsize±1.69 }&66.31{\scriptsize±2.10 }&57.93{\scriptsize±1.98 }&\textbf{67.99}{\scriptsize±1.63 }&58.48{\scriptsize±3.59 }&\textbf{\textcolor{black!40}{67.39}}{\scriptsize±3.12 }&65.19{\scriptsize±6.46 }\\ \cline{2-9}
&Diri(1.0)   &82.77{\scriptsize±0.59 }&82.10{\scriptsize±0.90 }&78.99{\scriptsize±1.76 }&82.34{\scriptsize±0.76 }&82.27{\scriptsize±0.78 }&\textbf{82.65}{\scriptsize±0.71 }&76.61{\scriptsize±2.68 }\\ \cline{2-9}
&LabelSkew(4)&71.88{\scriptsize±3.17 }&67.96{\scriptsize±0.48 }&62.71{\scriptsize±6.21 }&67.59{\scriptsize±1.75 }&67.26{\scriptsize±4.63 }&\textbf{73.97}{\scriptsize±3.64 }&\textbf{\textcolor{black!40}{72.31}}{\scriptsize±2.81 }\\ \cline{2-9}
 \toprule[1pt]
 \multirow{4}{*}{S-F} 
&IID       &81.60{\scriptsize±0.68 }&83.22{\scriptsize±0.63 }&82.71{\scriptsize±0.35 }&83.04{\scriptsize±0.73 }&\textbf{83.93}{\scriptsize±0.34 }&\textbf{\textcolor{black!40}{83.50}}{\scriptsize±0.82 }&83.48{\scriptsize±0.98 }\\ \cline{2-9}
&Diri(0.1)   &71.82{\scriptsize±1.69 }&63.01{\scriptsize±2.72 }&64.00{\scriptsize±3.28 }&63.67{\scriptsize±5.07 }&76.29{\scriptsize±0.70 }&\textbf{79.92}{\scriptsize±0.94 }&\textbf{\textcolor{black!40}{78.31}}{\scriptsize±1.05 }\\ \cline{2-9}
&Diri(1.0)   &82.77{\scriptsize±0.59 }&81.84{\scriptsize±1.00 }&80.15{\scriptsize±0.64 }&81.75{\scriptsize±0.50 }&\textbf{\textcolor{black!40}{82.96}}{\scriptsize±0.36 }&\textbf{83.34}{\scriptsize±0.23 }&82.50{\scriptsize±0.15 }\\ \cline{2-9}
&LabelSkew(4)&71.88{\scriptsize±3.17 }&62.55{\scriptsize±7.90 }&76.66{\scriptsize±4.77 }&71.34{\scriptsize±3.32 }&77.34{\scriptsize±2.17 }&\textbf{82.50}{\scriptsize±0.23 }&\textbf{\textcolor{black!40}{81.95}}{\scriptsize±0.30 }\\ \cline{2-9}
 \toprule[1pt]
 \multirow{4}{*}{ALIE} 
&IID       &81.60{\scriptsize±0.68 }&\textbf{83.79}{\scriptsize±0.20 }&82.71{\scriptsize±0.35 }&\textbf{\textcolor{black!40}{83.63}}{\scriptsize±0.60 }&83.59{\scriptsize±0.30 }&55.76{\scriptsize±4.89 }&83.49{\scriptsize±0.94 }\\ \cline{2-9}
&Diri(0.1)   &71.82{\scriptsize±1.69 }&72.54{\scriptsize±1.83 }&64.00{\scriptsize±3.28 }&69.83{\scriptsize±3.95 }&\textbf{\textcolor{black!40}{77.83}}{\scriptsize±1.66 }&\textbf{78.72}{\scriptsize±2.10 }&59.14{\scriptsize±4.00 }\\ \cline{2-9}
&Diri(1.0)   &82.77{\scriptsize±0.59 }&\textbf{\textcolor{black!40}{81.75}}{\scriptsize±0.87 }&80.15{\scriptsize±0.64 }&81.47{\scriptsize±1.73 }&\textbf{82.74}{\scriptsize±0.51 }&63.78{\scriptsize±8.45 }&80.70{\scriptsize±2.70 }\\ \cline{2-9}
&LabelSkew(4)&71.88{\scriptsize±3.17 }&69.18{\scriptsize±2.49 }&76.66{\scriptsize±4.77 }&76.15{\scriptsize±1.08 }&\textbf{\textcolor{black!40}{76.95}}{\scriptsize±1.43 }&30.71{\scriptsize±8.00 }&\textbf{81.97}{\scriptsize±0.21 }\\ \cline{2-9}
 \toprule[1pt]

    \end{tabular}

    }
    
    \caption{\textit{Byzantine attack robustness comparison (Acc{±std}) on MNIST and Fashion MNIST.} For BALANCE, hyperparameters are tuned to optimal performance ($\alpha = 0.1$, $\gamma = 2$ and $\kappa=1$).  For \textbf{DFedReweighting}, results are obtained using the TPM-CRS combination defined in \eqref{equ:combination-loss}. \textbf{Boldface} and \textbf{\textcolor{black!40}{grey}} denote the best and second-best accuracy within each scenario, excluding \textbf{No-AT}.}
    \label{tab:byzantine}
\end{table*}

\subsection{Experimental Results}
We conduct two sets of experiments to evaluate the proposed framework on client fairness and Byzantine robustness, followed by two illustrative examples demonstrating how to design TPM-CRS combinations for specific objectives.
\subsubsection{\textbf{Client Fairness}}
Client fairness in DFL aims to ensure comparable model performance across clients on their respective local test datasets. We quantify fairness using the variance of test accuracies across all $N$ clients, defined as $ \mathrm{Var} = \frac{1}{N}\sum_{k=1}^{N}\bigl(\mathrm{Acc}_{k}-\overline{\mathrm{Acc}}\bigr)^{2}$,
where $\mathrm{Acc}_k$ is the local test accuracy of client $k$ and $\overline{\mathrm{Acc}}$ is the mean accuracy across clients.
All clients train a softmax regression model with cross-entropy loss over $T = 3000$ rounds on  MNIST and Fashion-MNIST, with a fixed learning rate 0.01 a minibatch size of 32. We compare against FedAvg \cite{mcmahan2017communication} and $q$-FFL \cite{li2019fair} adapted to the DFL setting, referred to as DFedAvg and $q$-FDFL, respectively.

Since this is an image classification task, the local test accuracy serves as the TPM for client $k$. To promote fairness, the CRS adopts a temperature-scaled softmax reweighting strategy:  
\begin{equation}
    \begin{cases}
    m_k = \mathrm{Acc}_k  \\
    s(m_i, M_k) = \frac{e^{m_i/\mathcal{T}}}{\sum_{m_j \in M_k} e^{m_j/\mathcal{T}}},\; i\in\mathcal{N}_k
    \end{cases},
\label{equ:combination-fairness-acc}
\end{equation}
where a smaller temperature $\mathcal{T}$ yields a sharper, more concentrated weight distribution, while a larger $\mathcal{T}$ produces a softer, more uniform one. 

\textbf{Results.}
As shown in Table~\ref{tab:client_fairness-comparison}, under IID conditions, \textbf{DFedReweighting} achieves fairness comparable to the baselines while maintaining higher accuracy. Under non-IID conditions, it attains the highest accuracy while reducing variance to approximately 1\% to 50\%  of that observed in the baselines, depending on the specific scenario.

\subsubsection{\textbf{Byzantine Attack Robustness}}
Byzantine attack robustness in DFL aims to eliminate the adverse impact of malicious clients on the learning process~\cite{lamport2019byzantine}. We adopt the same image classification setup as in the client fairness experiments and simulate three representative Byzantine attacks: Gaussian Attack (GA)~\cite{fang2024byzantine,ye2023tradeoff}, which injects updates sampled from $\mathbb{N}(0, 30^2)$; Sign-Flipping Attack (S-F)~\cite{ye2023tradeoff}, which corrupts updates by multiplying local model weights by $-10$; and A Little Is Enough Attack (ALIE)~\cite{fang2024byzantine}, which constructs malicious updates using the mean and standard deviation of benign gradients.

In this setting, training loss is used as the TPM. To suppress the influence of Byzantine clients during aggregation, the CRS adopts a loss-clipping strategy that assigns negligible weight to updates with anomalously high loss:
\begin{equation}
    \begin{cases}
        m_k = f(\mathbf w_k, \mathcal{A}_k)  \\
        s(m_i, M_k) = \frac{m_i}{\sum_{m_j \in M_k} m_j}, 
        m_i =
        \begin{cases}
        m_i, & m_i \le \mu\\[4pt]
        0, & m_j > \mu
        \end{cases}, \; i\in\mathcal{N}_k
    \end{cases}
\label{equ:combination-loss}
\end{equation}
where $f(\cdot)$ denotes the training loss and $\mu$ is the mean loss across received updates. 

\textbf{Results.} 
We compare \textbf{DFedReweighting} against Median~\cite{yin2018byzantine}, mKrum, Trimmed Mean~\cite{yin2018byzantine}, Flame~\cite{nguyen2022flame}, and BALANCE~\cite{fang2024byzantine}. As shown in Table~\ref{tab:byzantine}, our method achieves the best performance in the majority of settings and remains competitive in the remainder. Notably, it occasionally surpasses the theoretical upper bounds of baseline methods, an effect we attribute to the regularizing influence of benign noise during training.

\begin{figure}[h]
    \centering
    \includegraphics[width=0.7\linewidth]{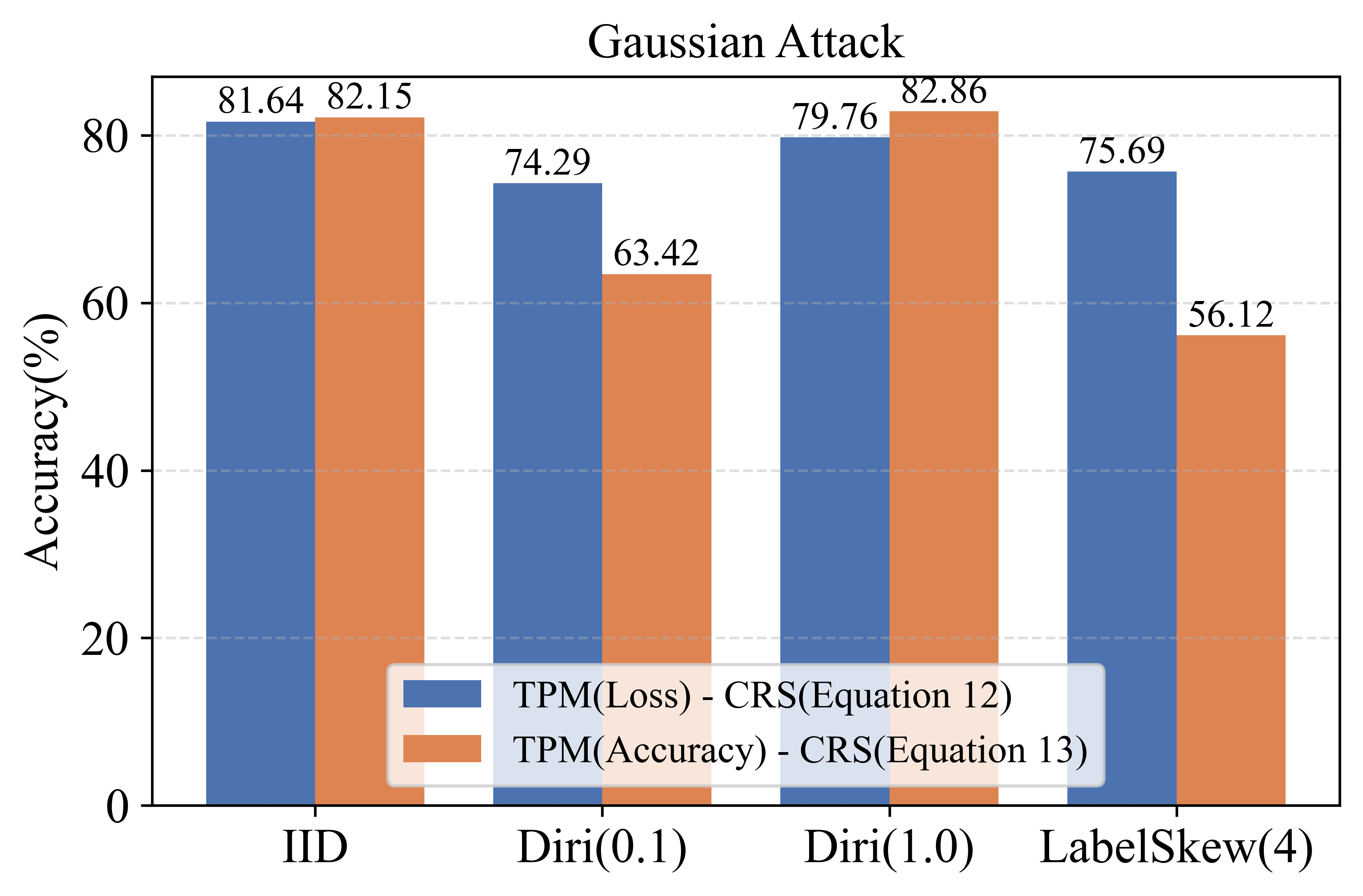}
    \caption{
  \textit{Impact of TPM-CRS combination choice on Byzantine attack robustness.} The loss-based combination (TPM: Loss, CRS: \eqref{equ:combination-loss}) consistently outperforms the accuracy-based alternative (TPM: Accuracy, CRS: \eqref{equ:strategy2inbzt}). Results under S-F and ALIE attacks follow the same trend.}

    \label{fig:compare}
\end{figure}

\subsubsection{\textbf{Another Alternative TPM-CRS Combination}} 
While the TPM-CRS design space is flexible, different combinations yield different performance outcomes. To illustrate this, we evaluate an alternative combination for Byzantine robustness in which accuracy replaces loss as the TPM, paired with an accuracy-based clipping rule as the CRS:
\begin{equation}
    \begin{cases}
        m_k = Acc_k \\
        s(m_i, M_k) = \frac{m_i}{\sum_{j \in M_k} m_j}, 
        m_i =
        \begin{cases}
        m_i, & m_i \ge \mu,\\[4pt]
        0, & m_i < \mu.
        \end{cases} 
    \end{cases},
\label{equ:strategy2inbzt}
\end{equation}
where $\mu$ is the mean accuracy serving as the clipping threshold. As shown in Fig.~\ref{fig:compare}, the loss-based combination \eqref{equ:combination-loss} consistently outperforms the accuracy-based alternative \eqref{equ:strategy2inbzt} under non-IID settings (Diri(0.1), Diri(1.0), and LabelSkew(4)), while achieving comparable performance under IID conditions, demonstrating its superior robustness to data heterogeneity.

\subsubsection{\textbf{Multi-objective Task across Clients}}
Multi-objective tasks across clients refer to settings where different clients pursue distinct learning objectives within the same DFL system. For instance, some clients may prioritize fairness while others seek to maximize accuracy. A natural solution is to equip each client with a different TPM-CRS combination aligned to its individual objective, as illustrated in Table~\ref{tab:multi-objects}.


\renewcommand{\arraystretch}{1.8} 
\begin{table}[!b]
\centering
\begin{tabular}{|l|c|c|c|}
\hline
Clients & Objective & TPM ($m_k$)& CRS ($s(m_i, M_k)$)\\ \hline
Client 1 & Client Fairness & $\mathrm{Acc}_k$  &  $\frac{e^{m_i/\mathcal{T}}}{\sum_{m_j \in M_k} e^{m_j/\mathcal{T}}}$ \\ \hline
Client 2 & Group Fairness & $\mathrm{Acc}_k$  &  $\frac{m_i}{\sum_{m_j \in M_k} m_j}$\\ \hline
Client 3 & Byzantine Robustness & $f(\mathbf w_k, \mathcal{A}_k)$  & $\frac{m_i}{\sum_{m_j \in M_k} m_j} $ \\ \hline
Client 4 & Fast Convergence & $f(\mathbf w_k, \mathcal{A}_k)$ & $\operatorname{clip}(m_i, 0, t)$\\ \hline
\end{tabular}
\caption{ Multi-objective task across clients in a four-client DFL system, readily scalable to larger networks and more diverse TPM-CRS combinations.}
\label{tab:multi-objects}
\end{table}

\subsubsection{\textbf{Multi-objective Task within Each Client}}
Multi-objective tasks within each client refer to settings where every client simultaneously optimizes multiple objectives~\cite{hu2022federated}, a scenario common in realistic DFL deployments. Unlike the across-client case, all clients here share the same multi-criteria learning goal and solve an identical multi-objective optimization problem. We thus propose the following TPM-CRS combination:
\begin{equation}
\small
    \begin{cases}
        m_k = \frac{1}{2}Acc_k + \frac{1}{2}f(\mathbf w_k, \mathcal{A}_k)\\
        s(m_i, M_k) = f_e
    \end{cases},
\label{equ:mult-within}
\end{equation}
where the TPM is a combination of local accuracy $Acc_k$ and training loss $f(\mathbf w_k, \mathcal{A}_k)$, and $f_e$ is a monotonically increasing function that assigns higher aggregation weights to updates with greater TPM scores.
\section{Conclusion}
This paper proposes DFedReweighting, a unified aggregation framework for decentralized federated learning that achieves diverse learning objectives via objective-oriented reweighting at the final step of each training round. The framework is broadly extensible: by appropriately designing the target performance metric (TPM) and customized reweighting strategy (CRS), it can be readily adapted to a wide range of DFL challenges. Theoretical analysis establishes linear convergence guarantees for general $L$-smooth and strongly convex functions under suitable TPM-CRS combinations. Empirical evaluations on client fairness and Byzantine robustness demonstrate consistent improvements over competitive baselines, and two multi-objective examples further illustrate the flexibility of the TPM-CRS design paradigm in accommodating diverse learning objectives.
\bibliographystyle{IEEEtran}
\bibliography{./sections/reference}
\end{document}